\documentclass{article}

\usepackage{microtype}
\usepackage{graphicx}
\usepackage{subcaption}
\usepackage{booktabs} 
\usepackage{arydshln} 
\usepackage{multirow} 
\usepackage{enumitem}
\usepackage{hyperref}

\usepackage[accepted]{icml2026}

\usepackage{amsmath}
\usepackage{amssymb}
\usepackage{mathtools}
\usepackage{amsthm}
\usepackage{xcolor}

\usepackage{algorithm}
\usepackage{algorithmic}

\usepackage[capitalize,noabbrev]{cleveref}

\theoremstyle{plain}

\theoremstyle{definition}

\theoremstyle{remark}

\DeclareMathOperator*{\argmax}{arg\,max}

\newcommand{\ars}{\textsc{AR w/ ret@static}}
\newcommand{\ara}{\textsc{AR w/ ret@adaptive}}
\newcommand{\arn}{\textsc{AR w/ ret@N}}
\newcommand{\arone}{\textsc{AR w/ ret@1}}
\newcommand{\arten}{\textsc{AR w/ ret@10}}
\newcommand{\dlms}{\textsc{DLM w/ ret@static}}
\newcommand{\dlmours}{\textsc{DLM w/ Sardi}}
\newcommand{\ours}{\textsc{Sardi}}

\usepackage{siunitx}
\usepackage{xcolor}
\definecolor{darkgreen}{RGB}{0,100,0}
\definecolor{BrickRed}{rgb}{0.8, 0.25, 0.33}

\icmltitlerunning{Self-Augmenting Retrieval for Diffusion Language Models}

\begin{document}

\twocolumn[
  \icmltitle{Self-Augmenting Retrieval for Diffusion Language Models}



  \icmlsetsymbol{equal}{*}

  \begin{icmlauthorlist}
    \icmlauthor{Paul J\"unger}{cornell}
    \icmlauthor{Justin Lovelace}{cornell}
    \icmlauthor{Linxi Zhao}{cornell}
    \icmlauthor{Dongyoung Go}{cornell}
    \icmlauthor{Kilian Q. Weinberger}{cornell}
  \end{icmlauthorlist}

  \icmlaffiliation{cornell}{Department of Computer Science, Cornell University}

  \icmlcorrespondingauthor{Paul J\"unger}{pj287@cornell.edu}

  \icmlkeywords{diffusion language models, discrete diffusion, non-autoregressive generation, retrieval, iterative refinement}

  \vskip 0.3in
]



\printAffiliationsAndNotice{}  

\begin{abstract}
Diffusion language models generate text by iteratively denoising an entire response in parallel, holding a noisy draft of the answer at every step. We show that these intermediate states surface salient entities early in the denoising trajectory, making them a useful lookahead signal for retrieval-augmented generation (RAG).
We further demonstrate that grounding generation in retrieved evidence sharply reduces inter-token dependence, making RAG particularly well suited to parallel decoding. We exploit both through Self-Augmenting Retrieval for Diffusion Language Models (SARDI), a dynamic RAG framework that uses speculative token predictions to guide retrieval during denoising. SARDI is training-free, retriever-agnostic, and applicable to any reasoning-capable discrete diffusion language model. Across five multi-hop QA benchmarks, SARDI outperforms training-free diffusion and autoregressive retrieval baselines at up to $4\times$ higher throughput. Our code is available at \url{https://github.com/pauljngr/SARDI}.
\end{abstract}

\section{Introduction}
\label{sec:intro}

Large language models are increasingly deployed as interfaces to external knowledge, with retrieval-augmented generation grounding responses in retrieved evidence~\cite{lewis2020rag, izacard2021fid}. Yet the dominant decoding paradigm, token-by-token autoregressive (AR) generation, ties latency directly to output length~\cite{pope2023_inference} and forces retrieval to condition on a left-to-right committed prefix~\cite{trivedi2023interleaving}.
\emph{Diffusion language models} (DLMs) offer an alternative. Rather than emitting one token at a time, they iteratively denoise a corrupted sequence, updating many positions in parallel at each step with a test-time-adjustable number of iterations~\cite{austin2021d3pm, li2022diffusionlm, snell2024scalingllmtesttimecompute, dream7b, lovelace2023latent}. Recent discrete DLMs such as DREAM-7B~\cite{dream7b} and LLaDA~\cite{nie2025llada} have matured into competitive models at scale, raising the question of how their non-autoregressive structure can be exploited beyond raw throughput.

Denoising unlocks two structural advantages over AR generation in retrieval-augmented settings: one for retrieval and one for decoding.

\begin{figure*}[t]
  \centering
  \includegraphics[width=1\textwidth]{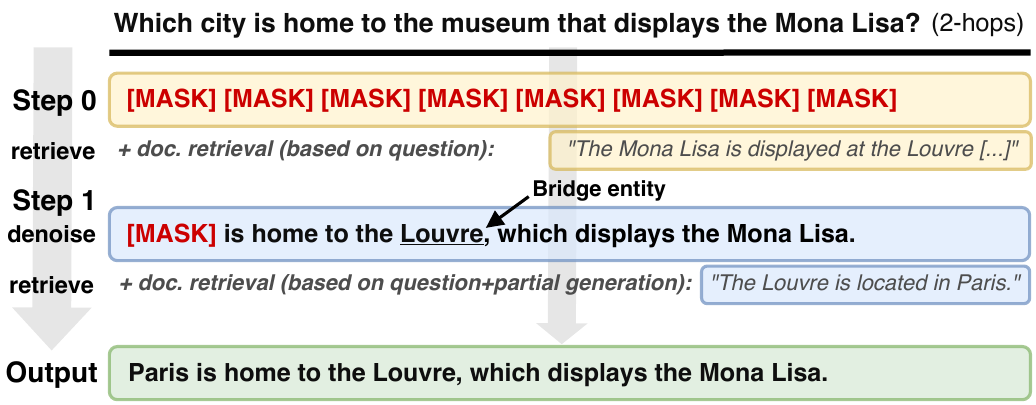}
  \caption{Static question-only retrieval can fail on multi-hop QA when the question does not specify the \underline{bridge entity}. Intermediate diffusion states often surface such entities early, enabling retrieval of later-hop evidence before the output is finalized.}
  \label{fig:motivation}
\end{figure*}

\paragraph{Diffusion Trajectories as Lookahead for Retrieval.} Diffusion language models refine the entire response at once: at every denoising step they produce tentative predictions for \emph{all} token positions. This trajectory of intermediate predictions surfaces salient entities and relations before the output is finalized. This is especially useful for multi-hop question answering, where the evidence needed for later reasoning steps depends on intermediate \emph{bridge} entities that the question alone does not specify. \cref{fig:motivation} illustrates this: asked which city is home to the museum that displays the Mona Lisa, a static retriever cannot retrieve the passage \emph{``The Louvre is located in Paris''} without first identifying the bridge entity \emph{Louvre}. Intermediate diffusion states surface this entity early in the trajectory, enabling retrieval of the second-hop evidence before the final answer is committed. Diffusion trajectories thus let the model ``peek into the future'' of its own generation to improve retrieval.

\begin{figure}
  \centering
  \includegraphics[width=1\columnwidth]{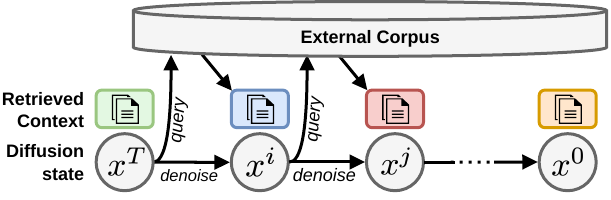}
  \caption{Diffusion Language Models hold a noisy draft of the entire response at every denoising step. SARDI turns that draft into a retrieval query, retrieves new documents, and conditions the next step on the refreshed evidence. This way, evidence needed for later reasoning hops can be retrieved as the response takes shape.}
\label{fig:retrieve_during_denoising}
\end{figure}

\paragraph{RAG grounding promotes parallel decoding.}
Realizing the speedup potential of parallel decoding has proven challenging in practice: sampling multiple tokens simultaneously risks generating conflicting or incoherent spans, errors that can cascade through the response and degrade output quality~\cite{wu2026fastdllm}. RAG changes this. When generation is grounded in informative evidence $D$, many output tokens are copied or paraphrased straight from the context and are thus conditionally independent given $D$. We confirm this empirically in \cref{sec:cmi}: grounding sharply reduces inter-token dependence, promoting parallel decoding.

We propose \textbf{Self-Augmenting Retrieval for Diffusion Language Models (\ours)}, the first framework to condition retrieval on intermediate diffusion states. \ours\ interleaves retrieval with denoising: at each iteration, it constructs a query from the partially denoised sequence, retrieves fresh evidence, and conditions the next step on the updated context (Figure~\ref{fig:retrieve_during_denoising}). Central to \ours\ is a \textbf{separation between retrieval and generation confidence} unique to non-autoregressive decoders: speculative future tokens can inform retrieval {long before} they are stable enough to commit to the output. \ours\ is training-free and works plug-and-play with any discrete diffusion language model that can produce reasoning traces.

We empirically validate two properties that make diffusion language models well suited to retrieval. First, we demonstrate that intermediate denoising states surface bridge entities earlier than autoregressive generation, providing a stronger signal for retrieval. Second, grounding generation in retrieved evidence sharply reduces inter-token mutual information, making the response easier to decode in parallel. Together, this lets \ours\ dominate the quality--latency frontier across five multi-hop QA benchmarks, outperforming current training-free diffusion and autoregressive baselines at substantially lower latency.

\section{Related Work}
\label{sec:related}

\paragraph{Retrieval-augmented generation.}
Early RAG systems~\citep{lewis2020rag, izacard2021fid, karpukhin2020dpr} perform \emph{single-shot} retrieval from the input query and keep the retrieved context fixed throughout generation. This constrains their effectiveness on multi-hop tasks, where the evidence needed for later reasoning steps depends on bridge entities that the question does not name~\citep{yang2018hotpotqa, ho-etal-2020-2wiki}. All existing RAG approaches for diffusion language models are restricted to this single-shot paradigm~\citep{yu2026unlocking, fang_commonsense}.

\begin{figure*}[t]
  \centering
  \includegraphics[width=\textwidth]{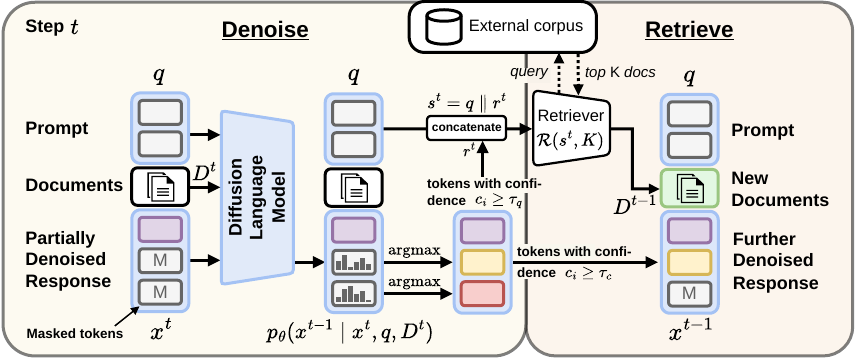}
  \caption{Overview of self-augmenting retrieval for diffusion language models. At step $t$, the diffusion LM denoises the partially masked response. Tokens with confidence $c_i \ge \tau_q$ then form a query to refresh the retrieved evidence, while only the more confident tokens ($c_i \ge \tau_c$) are committed to the next response $x^{t-1}$. Speculative tokens can thus inform retrieval before they are stable enough to commit.}
  \label{fig:method-overview}
\end{figure*}

\paragraph{Dynamic and agentic retrieval in autoregressive LMs.}
For autoregressive models, this single-shot limitation has been addressed by a line of work that interleaves retrieval with generation, deriving follow-up queries from the tokens produced so far~\citep{trivedi2023interleaving, jiang2023active, jeong2024adaptive, asai2023selfrag}.
Most relevant to this work is FLARE~\citep{jiang2023active}, which makes retrieval \emph{forward-looking}: rather than querying only on already-committed text, FLARE first generates a tentative next sentence and, if that sentence contains low-confidence tokens, uses it as a query to retrieve fresh evidence. Then, it regenerates the sentence under the updated context. While anticipating future tokens does help, autoregressive decoding makes this lookahead fragile: the tentative span is generated left-to-right, so a single early error can compound and produce hallucinated queries that retrieve irrelevant documents (see results in Section~\ref{sec:main-results}). In contrast, \ours{} predicts all tentative tokens in parallel, so errors do not compound.

More recently, agentic systems generate explicit search queries via planning and self-reflection~\citep{yao2022react, asai2023selfrag, xu2024activerag, li2025searcho1, jin2025searchr1}. While effective, these approaches typically require specialized training via reinforcement learning, increasing engineering complexity and computational overhead. In contrast, \ours{} is \emph{plug-and-play} and requires no learned retrieval controller or query generator. To the best of our knowledge, \ours{} is the first retrieval framework that conditions retrieval on intermediate diffusion states and refreshes evidence throughout the denoising trajectory.

\section{Background}
\label{sec:background}

We consider open-domain (multi-hop) question answering. Given a question $q$, the goal is to generate a response $x = (x_1, \dots, x_L) \in \mathcal{V}^L$ of length $L$ over a finite vocabulary $\mathcal{V}$. Retrieval-augmented generation (RAG) retrieves $K$ passages $D = \{d_1, \dots, d_K\}$ from an external corpus $\mathcal{C}$ via a query $s$, and conditions generation on both $q$ and $D$.

\subsection{Autoregressive Language Models}
Autoregressive (AR) language models factor the response likelihood left-to-right:
\begin{equation}
  p_\theta(x \mid q, D) = \prod_{i=1}^{L} p_\theta(x_i \mid x_{<i}, q, D),
  \label{eq:ar-factorization}
\end{equation}
where $x_{<i} = (x_1, \dots, x_{i-1})$ and $\theta$ denotes the model parameters. Decoding proceeds one token at a time, so retrieval can only condition on the committed prefix $x_{<i}$.

\subsection{Discrete Diffusion Language Models}
\label{sec:dlm-background}
Discrete diffusion language models (DLMs) process all positions in parallel through iterative denoising. A DLM defines a sequence of states $\{x^t\}_{t=0}^T$, where $x^T = ([\texttt{MASK}], \dots, [\texttt{MASK}])$ is fully masked and $x^0 \in \mathcal{V}^L$ is the final output. Generation proceeds from $t = T$ to $t = 0$: at each step, given a partially masked sequence $x^t \in (\mathcal{V} \cup \{[\texttt{MASK}]\})^L$, $q$, and $D$, the denoiser predicts a per-position distribution
\begin{equation}
  p_\theta(x^{t-1}_{i} \mid x^t, q, D), \qquad i \in \{1, \dots, L\},
  \label{eq:dlm-denoiser}
\end{equation}
and selected masked positions are \emph{unmasked} to their argmax.

\paragraph{Parallel decoding and RAG.}
Sampling positions independently approximates the joint distribution $p_\theta(x \mid q, D)$ by the product of its marginals. This breaks down under strong inter-token dependence. For example, the first name ``Albert'' makes ``Einstein'' very likely, whereas independent sampling can produce ``Albert Newton''. We argue that RAG sharply reduces such dependence: tokens copied or paraphrased from retrieved passages are largely determined by $D$, leaving adjacent positions nearly independent. We validate this hypothesis in \cref{sec:cmi}, confirming that RAG is a regime well-suited to parallel decoding.

\section{Self-Augmenting Retrieval for Diffusion}
\label{sec:method}

Diffusion Language Models hold a noisy draft of the entire answer at every denoising step: 
the diffusion trajectory $\{x^t\}_{t=0}^T$ thus exposes a sequence of intermediate states at which retrieval can be revisited. Whereas static RAG issues a single query from the question, \emph{Self-Augmenting Retrieval for Diffusion} (\ours) refines the query as the response takes shape, surfacing new evidence at every step. \ours\ interleaves retrieval with denoising (\cref{fig:method-overview}). At each step $t$, the model predicts a token for every masked position with a confidence score $c_i = \max_{v \in \mathcal{V}} p_\theta(v \mid x^t, q, D^t)$. Depending on $c_i$, a position is used to:
\begin{enumerate}[label=(\roman*),leftmargin=*,itemsep=2pt,topsep=2pt]
  \item \textbf{Retrieve.} If $c_i \ge \tau_q$ (the \emph{query threshold}), the predicted token is added to the retrieval query, refreshing the evidence for the next step.
  \item \textbf{Commit.} If $c_i \ge \tau_c$ (the \emph{commit threshold}), the token prediction is committed to; the rest of the tokens are remasked.
\end{enumerate}
Separating the two thresholds is the central design choice in \ours: because $\tau_q \le \tau_c$, tentative tokens can inform retrieval well before they are reliable enough to commit. The following sections detail query construction (\cref{sec:query}), evidence refresh (\cref{sec:refresh}), and confidence-based commitment (\cref{sec:unmasking}).

\subsection{Query Construction}
\label{sec:query}
As denoising progresses, the emerging response surfaces intermediate entities --- names, dates, relations --- that the question does not contain but that later-hop retrieval needs. We want to feed these to the retriever as early as possible, even while they are still uncertain. This is safe because retrieval and generation tolerate errors very differently: committing an incorrect token can directly corrupt the output, whereas retrieval is robust to noisy queries. We can therefore set the query threshold $\tau_q$ well below the commit threshold $\tau_c$, exposing tentative tokens to the retriever long before they are reliable enough to commit. At $\tau_q = 0$ every position enters the query (maximal lookahead); raising $\tau_q$ restricts it to more confident tokens.

Concretely, let $x^t \in (\mathcal{V} \cup \{\texttt{[MASK]}\})^L$ be the current sequence and define $c_i = \max_{v \in \mathcal{V}} p_\theta(v \mid x^t, q, D^t)$ as the model confidence at position $i$. We form the proxy sequence $\tilde{x}^t$ by using token predictions with confidence at least $\tau_q$:
\begin{equation}
  \tilde{x}^t_i =
  \begin{cases}
    x^t_{i}, & x^t_{i} \neq \texttt{[MASK]} \\
    \argmax_{v} p_\theta(v \mid x^t, q, D^t), & c_i \geq \tau_q \\
    \texttt{[MASK]}, & \text{otherwise.}
  \end{cases}
  \label{eq:proxy}
\end{equation}
The proxy is detokenized to $r^t = \mathrm{Detokenize}(\tilde{x}^t)$ (remaining $[\texttt{MASK}]$ tokens are dropped), and the retrieval query concatenates the question with this intermediate response:
\begin{equation}
  s^t = q \;\|\; r^t.
  \label{eq:query}
\end{equation}
The fixed question $q$ anchors the query when early predictions are noisy, while the evolving $r^t$ progressively specializes retrieval. Unless otherwise specified we use $\tau_q = 0$; \cref{sec:lookahead} reports an empirical sweep validating this choice.

\begin{algorithm}[t]
\caption{\ours: Self-Augmenting Retrieval for Diffusion Language Models}
\label{alg:sar}
\begin{algorithmic}[1]
\REQUIRE Question $q$, retriever $\mathcal{R}$, denoiser $p_\theta$, query threshold $\tau_q$, commit threshold $\tau_c$, context size $K$
\ENSURE Generated sequence $x$
\STATE $x \leftarrow (\texttt{[MASK]}, \dots, \texttt{[MASK]})$ \COMMENT{Fully masked initialization}
\STATE $D \leftarrow \mathcal{R}(q, K)$ \COMMENT{Initial retrieval from question only}
\WHILE{not fully unmasked}
\vspace{2mm}
    \STATE \textit{// --- Decode: commit high-confidence tokens ---}
    \STATE Compute $p_\theta(\cdot \mid x, q, D)$ for all masked positions
    \STATE $c_i \leftarrow \max_{v \in \mathcal{V}} p_\theta(v \mid x, q, D)$ for each masked position $i$
    \STATE $\mathcal{U} \leftarrow \{ i : x_i = \texttt{[MASK]} \wedge c_i \ge \tau_c \}$
    \IF{$\mathcal{U} = \emptyset$}
        \STATE $\mathcal{U} \leftarrow \left\{\argmax_{i: x_i=\texttt{[MASK]}} c_i\right\}$ \COMMENT{Ensure progress}
    \ENDIF
    \STATE For all $i \in \mathcal{U}$: \; $x_i \leftarrow \argmax_{v \in \mathcal{V}} p_\theta(v \mid x, q, D)$
    \vspace{2mm}
    \STATE \textit{// --- Retrieve: refresh evidence ---}
    \STATE Construct proxy $\tilde{x}$ by filling masks with $c_i \ge \tau_q$ via argmax (\cref{eq:proxy})
    \STATE $s \leftarrow q \,\|\, \mathrm{Detokenize}(\tilde{x})$ \COMMENT{Form retrieval query}
    \STATE $D \leftarrow \mathcal{R}(s, K)$ \COMMENT{Retrieve new evidence}
\ENDWHILE
\STATE \textbf{return} $x$
\end{algorithmic}
\end{algorithm}

\subsection{Evidence Refresh}
\label{sec:refresh}

Using the query $s^t$ constructed above, \ours\ retrieves a fresh set of $K$ passages at each step:
\begin{equation}
  D^{t-1} \leftarrow \mathcal{R}(s^t, K),
  \label{eq:retrieve}
\end{equation}
where $\mathcal{R}$ is the retriever. The new evidence $D^{t-1}$ replaces the previous context entirely and conditions the next denoising step. We use BM25~\citep{bm25} in our experiments for efficiency, but \ours\ is retriever-agnostic and works with any sparse or dense retriever.

\subsection{Confidence-Based Unmasking}
\label{sec:unmasking}

Because \ours\ refreshes evidence at every step, the order in which tokens are committed directly shapes downstream retrieval. We adopt threshold-based unmasking~\citep{wu2026fastdllm}, which reveals all positions whose confidence exceeds the commit threshold $\tau_c$:
\begin{equation}
  \mathcal{U}^t = \bigl\{ i \mid x^t_{i} = \texttt{[MASK]} \;\wedge\; c_i \geq \tau_c \bigr\},
  \label{eq:unmask-set}
\end{equation}
where $c_i = \max_{v \in \mathcal{V}} p_\theta(v \mid x^t, q, D^t)$.

Each unmasked position commits to its argmax:
\begin{equation}
  x^{t-1}_{i} \leftarrow \argmax_{v \in \mathcal{V}} \, p_\theta(v \mid x^t, q, D^t).
  \label{eq:argmax-decode}
\end{equation}
If no position exceeds $\tau_c$, the single most confident masked position is unmasked to guarantee progress.

This creates a natural curriculum: high-confidence tokens (often text spans that can already be grounded in the current evidence) commit first and inform the next retrieval, while uncertain spans wait for refined evidence. The commit threshold $\tau_c$ therefore controls both decoding parallelism and the granularity of retrieval refinement, offering a single knob to trade accuracy against throughput (\cref{fig:accuracy_vs_qps}). \Cref{alg:sar} gives the full procedure.

\begin{table*}[t]
  \caption{Main results on multi-hop QA benchmarks (Exact Match $\times 100$).
  $\pm$~denotes bootstrap standard deviation; Time is mean wall-clock seconds per example. CofCA and SynthWorlds-SM use \emph{counterfactual} corpora (made-up facts) to prevent data-leakage. Search-R1 is grayed as it requires additional RL training (see \cref{sec:search-agents}).}
  \label{tab:main-results}
  \centering
  \renewcommand{\arraystretch}{1.15}
  \resizebox{\textwidth}{!}{%
    \begin{tabular}{lcccccccccc}
      \toprule
      & \multicolumn{2}{c}{\textbf{2WikiMultihopQA}}
      & \multicolumn{2}{c}{\textbf{HotpotQA}}
      & \multicolumn{2}{c}{\textbf{CofCA}}
      & \multicolumn{2}{c}{\textbf{MuSiQue}}
      & \multicolumn{2}{c}{\textbf{SynthWorlds-SM}} \\
      \cmidrule(lr){2-3} \cmidrule(lr){4-5} \cmidrule(lr){6-7} \cmidrule(lr){8-9} \cmidrule(lr){10-11}
      \textbf{Method}
        & \textbf{EM} & \textbf{Time}
        & \textbf{EM} & \textbf{Time}
        & \textbf{EM} & \textbf{Time}
        & \textbf{EM} & \textbf{Time}
        & \textbf{EM} & \textbf{Time} \\
      \midrule
      \textit{Training-free; autoregressive}\\
      \ars
        & 44.5\scriptsize{$\pm$0.6} & 0.74
        & 40.3\scriptsize{$\pm$0.8} & 0.66
        & 41.0\scriptsize{$\pm$1.6} & 0.76
        & 12.7\scriptsize{$\pm$0.7} & 0.69
        & 16.5\scriptsize{$\pm$1.1} & 0.81 \\
      \arten
        & 53.6\scriptsize{$\pm$0.6} & 0.80
        & 45.1\scriptsize{$\pm$0.8} & 0.75
        & 39.6\scriptsize{$\pm$1.6} & 0.83
        & 17.7\scriptsize{$\pm$0.8} & 0.74
        & 19.3\scriptsize{$\pm$1.1} & 0.95 \\
      \arone
        & 58.8\scriptsize{$\pm$0.6} & 1.26
        & 47.4\scriptsize{$\pm$0.8} & 1.40
        & 41.2\scriptsize{$\pm$1.6} & 1.37
        & 19.8\scriptsize{$\pm$0.8} & 1.23
        & 20.4\scriptsize{$\pm$1.1} & 2.08 \\
      \ara
        & 46.6\scriptsize{$\pm$0.6} & 1.47
        & 37.3\scriptsize{$\pm$0.8} & 1.34
        & 32.4\scriptsize{$\pm$1.6} & 1.49
        & 17.0\scriptsize{$\pm$0.8} & 1.42
        & 15.7\scriptsize{$\pm$1.1} & 1.71 \\
      AdaptiveRAG~\citep{jeong2024adaptive}
        & 34.1\scriptsize{$\pm$0.6} & 5.37
        & 37.1\scriptsize{$\pm$0.8} & 4.07
        & 38.4\scriptsize{$\pm$1.6} & 2.88
        & 14.9\scriptsize{$\pm$0.7} & 7.38
        & 13.2\scriptsize{$\pm$1.0} & 8.94 \\
      ReAct~\citep{yao2022react}
        & 42.7\scriptsize{$\pm$0.6} & 2.15
        & 40.1\scriptsize{$\pm$0.8} & 2.07
        & 42.9\scriptsize{$\pm$1.6} & 2.02
        & 20.9\scriptsize{$\pm$0.8} & 2.32
        & 22.2\scriptsize{$\pm$1.2} & 3.01 \\
      \midrule
      \textit{Training-free;  diffusion}\\
      {$\dlms$ $\tau_c{=}0.9$}
        & 43.7\scriptsize{$\pm$0.6} & 0.46
        & 39.9\scriptsize{$\pm$0.8} & 0.55
        & 43.4\scriptsize{$\pm$1.6} & 0.79
        & 11.1\scriptsize{$\pm$0.7} & 1.01
        & 14.4\scriptsize{$\pm$1.0} & 1.49 \\
      {$\dlmours$ (Ours) $\tau_c{=}0.9$}
        & 57.8\scriptsize{$\pm$0.6} & 0.39
        & 48.5\scriptsize{$\pm$0.8} & 0.64
        & 45.3\scriptsize{$\pm$1.6} & 0.75
        & 20.5\scriptsize{$\pm$0.8} & 0.88
        & 21.1\scriptsize{$\pm$1.2} & 1.29 \\
      {$\dlmours$ (Ours) $\tau_c{=}0.95$}
        & 59.1\scriptsize{$\pm$0.6} & 0.56
        & 48.7\scriptsize{$\pm$0.8} & 0.90
        & 44.9\scriptsize{$\pm$1.6} & 1.09
        & 20.6\scriptsize{$\pm$0.8} & 1.19
        & 21.7\scriptsize{$\pm$1.2} & 1.78 \\
      \midrule
      \textit{RL-trained; autoregressive}\\
      \textcolor{gray}{AR (Search-R1)~\citep{jin2025searchr1}}
        & \textcolor{gray}{52.4\scriptsize{$\pm$0.6}} & \textcolor{gray}{3.36}
        & \textcolor{gray}{50.3\scriptsize{$\pm$0.8}} & \textcolor{gray}{2.83}
        & \textcolor{gray}{44.4\scriptsize{$\pm$1.6}} & \textcolor{gray}{3.00}
        & \textcolor{gray}{26.4\scriptsize{$\pm$0.9}} & \textcolor{gray}{3.14}
        & \textcolor{gray}{26.9\scriptsize{$\pm$1.3}} & \textcolor{gray}{3.74} \\
      \bottomrule
    \end{tabular}
  }
\end{table*}

\section{Experiments}
\label{sec:experiments}

\subsection{Experimental Setup}
\label{sec:exp}

\paragraph{Datasets and benchmarks.}
We evaluate \ours\ on five multi-hop QA benchmarks: 2WikiMultiHopQA~\cite{ho-etal-2020-2wiki}, HotpotQA~\cite{yang2018hotpotqa}, MuSiQue~\cite{trivedi-etal-2022-musique}, CofCA~\cite{wu2024cofcastepwisecounterfactualmultihop}, and SynthWorlds-SM~\cite{gu2025synthworldscontrolledparallelworlds}, reporting Exact Match (EM) following~\citet{jin2025searchr1}. To measure per-query retrieval recall, the corpus must be at the same passage granularity as each benchmark's gold annotations: HotpotQA provides a full Wikipedia corpus at this granularity~\cite{yang2018hotpotqa}, and for the other four we follow prior work~\citep{prabhu2024dexterbenchmarkopendomaincomplex, yu2026unlocking} and concatenate all gold and distractor passages into a single corpus. Unless stated otherwise we use the sparse retrieval method BM25~\cite{bm25}, retrieving $K{=}7$ passages per iteration, and additionally evaluate the \emph{E5-base-v2} dense retriever in~\cref{sec:ablations}. Latency is measured in wall-clock time unbatched on a single NVIDIA B200 GPU.

\paragraph{RAG and reasoning with DLMs.}
\label{sec:exp_sft}
\ours\ relies on the existence of a reasoning trace in order to surface bridge entities. In our analysis, off-the-shelf instruction-tuned diffusion LMs such as \textsc{DREAM-7B} almost never produce reasoning traces in a RAG setting, even when using few-shot prompting (\cref{tab:sft-failure}, App.~\ref{appendix:sft-analysis}). \citet{trivedi2023interleaving} made an analogous observation for AR models in 2023:
``IRCoT relies on the base LM to have a zero or few-shot CoT-generation ability [\dots] not as common for small LMs (under 20B) [\dots] smaller LMs will likely increasingly acquire such ability.''~\cite{trivedi2023interleaving} Their prediction proved correct for AR models, and we expect the same for DLMs as they mature.

For now, to evaluate our retrieval mechanism fairly, we elicit the model to output reasoning via light supervised fine-tuning on synthetic chain-of-thought traces from \emph{gpt-4o-mini} (App.~\ref{appendix:sft-analysis}). To ensure a fair comparison, we apply \emph{identical} fine-tuning to \textsc{DREAM-7B} and the AR model Qwen2.5-7B. Crucially, after fine-tuning the two models reach near-identical EM under static question-only retrieval ($K{=}7$) on 2WikiMultiHopQA: $\dlms$ achieves $43.7\%$ and $\ars$ achieves $44.5\%$ (Table~\ref{tab:main-results}). To ensure the gains do not come from memorization, we include two counterfactual benchmarks. CofCA and SynthWorlds-SM use corpora with made-up facts that are absent from both pre-training and our fine-tuning data. Finally, we emphasize that \ours\ is not tied to the specific reasoning format induced by the synthetic \emph{gpt-4o-mini} traces; any reasoning style that includes intermediate entities will work.

\subsection{Baselines}
\label{sec:baselines}

We compare \ours\ against retrieval-augmented baselines from both AR and diffusion paradigms. AR baselines use Qwen2.5-7B and diffusion baselines use \textsc{DREAM-7B}, with the same light fine-tuning (\cref{sec:exp_sft}) applied to both backbones. The one exception is Search-R1, which we evaluate from the authors' released RL-trained checkpoint (App.~\ref{appendix:searchr1-setup}). All methods share identical prompts and evidence formatting (App.~\ref{appendix:prompts}), isolating the retrieval mechanism.

\begin{figure*}[th]
  \centering
  \includegraphics[width=\textwidth]{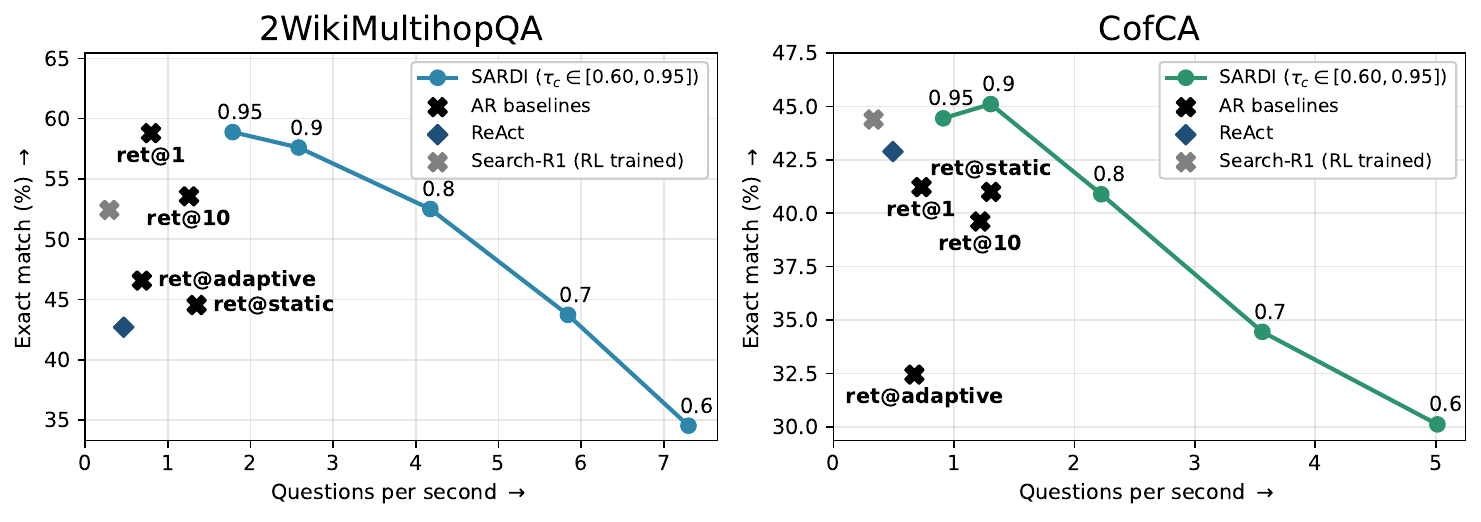}
  \caption{Accuracy vs.\ throughput trade-off on 2WikiMultiHopQA and CofCA, traced by adjusting the commit threshold $\tau_c$. Similar pareto patterns hold on HotpotQA, MuSiQue, and SynthWorlds-SM (\cref{fig:accuracy_vs_qps-appendix}).}
  \label{fig:accuracy_vs_qps}
\end{figure*}

\begin{itemize}[leftmargin=*, itemsep=2pt, topsep=0pt]
    \item \textbf{\ars}: Single retrieval from $q$; $D$ is fixed throughout generation.
    \item \textbf{\arn}: Retrieval every $N \in \{1, 10\}$ tokens.
    \item \textbf{\ara}~\cite{jiang2023active} (FLARE): Confidence-triggered retrieval using low-confidence spans as both triggers and queries.
    \item \textbf{AdaptiveRAG}~\citep{jeong2024adaptive}: AR LM equipped with a query-complexity router that triggers either single-step or iterative retrieval.
    \item \textbf{ReAct}~\citep{yao2022react}: A training-free agentic loop with \texttt{Retrieve[query]} and \texttt{Finish[ans]} actions.
    \item \textbf{\dlms}: The diffusion counterpart of \ars; one retrieval, fixed $D$.
    \item \textbf{\dlmours} (this work): \textsc{DREAM-7B} equipped with \ours.
    \item \textbf{AR (Search-R1)}~\citep{jin2025searchr1}: An RL-trained search agent that learns to emit explicit search queries during generation. Included as a strong reference point (discussed in \cref{sec:main-results}).
\end{itemize}

\subsection{\ours\ Is Faster and More Accurate}
\label{sec:main-results}

\paragraph{Accuracy and throughput.}
\ours\ improves substantially over static diffusion retrieval on every benchmark (\cref{tab:main-results}): on 2WikiMultiHopQA it raises EM from $44$ to $59$, with similar gains on HotpotQA ($40\to49$), CofCA ($43\to45$), MuSiQue ($11\to21$), and SynthWorlds-SM ($14\to22$). It also matches or beats every training-free AR baseline, and does so at much lower latency. We can trade off throughput against accuracy via the commit confidence threshold $\tau_c$: a higher $\tau_c$ commits fewer tokens per denoising step and triggers more retrieval rounds (higher accuracy, lower throughput), while a lower $\tau_c$ does the reverse. Sweeping $\tau_c$ (\cref{fig:accuracy_vs_qps}), \ours\ dominates the quality--latency frontier, running up to $4\times$ faster than AR iterative-retrieval baselines at comparable or better accuracy. We analyze the sources of these gains in the following sections.

\paragraph{Comparison to trained search agents.}\label{sec:search-agents}
Search-R1~\citep{jin2025searchr1} is an RL-trained search agent that learns to emit explicit search queries during generation.
We include it as a strong reference point but emphasize that \textbf{\ours\ and Search-R1 occupy different design points}: Search-R1 invests in RL-based query generation to maximize accuracy at inference cost, whereas \ours\ is a training-free, plug-and-play retrieval mechanism that requires no reward design or policy optimization.
Across the five benchmarks \ours\ is the strongest training-free method while running $4$--$9\times$ faster than Search-R1 at comparable accuracy on 2WikiMultiHopQA, HotpotQA, and CofCA (\cref{tab:main-results}).
The two approaches are complementary rather than competing, and combining diffusion trajectories with light RL-based query supervision is a natural direction for future work.

\subsection{Diffusion Trajectories as Lookahead for Retrieval}
\label{sec:lookahead}

A central hypothesis of this work is that the uncommitted, low-confidence tokens in the diffusion trajectory are a useful lookahead signal for retrieval. Because a DLM denoises the whole sequence at once, it holds tentative predictions for every position long before they are committed. \ours\ feeds these speculative tokens to the retriever, surfacing later-hop evidence  early on in the generation process. The following experiments test this hypothesis.

\begin{figure}[t]
  \centering
  \includegraphics[width=\columnwidth]{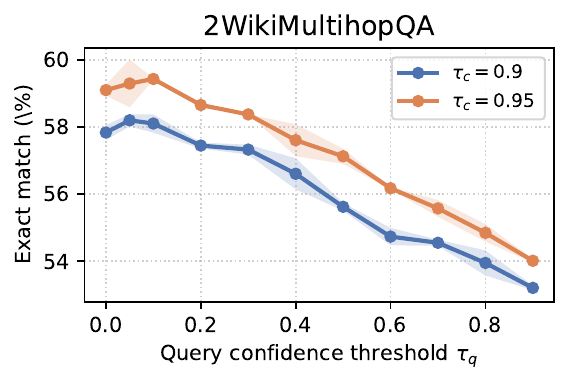}
  \caption{Sweep of the query threshold $\tau_q$ on 2WikiMultiHopQA (BM25, $K{=}7$), at the two commit thresholds $\tau_c \in \{0.9, 0.95\}$. Similar trend holds on HotpotQA, MuSiQue, and SynthWorlds-SM (\cref{tab:qct-sweep}).}
  \label{fig:qct-sweep}
\end{figure}

\paragraph{More lookahead is better.}
The query threshold $\tau_q$ sets how confident the model must be in a token before exposing it to the retriever (\cref{eq:proxy}): $\tau_q{=}0$ exposes every position, while $\tau_q{=}\tau_c$ restricts the query to tokens that are confident enough to be committed. Sweeping $\tau_q$ on 2WikiMultiHopQA (\cref{fig:qct-sweep}), EM peaks at the most aggressive setting ($\tau_q \approx 0$) and falls as the query becomes more conservative; similar trends hold on HotpotQA, MuSiQue, and SynthWorlds-SM (\cref{tab:qct-sweep}). This directly supports our hypothesis that speculative future tokens can inform retrieval long before they are stable enough to commit to the output.

\begin{figure}[t]
  \centering
  \includegraphics[width=\columnwidth]{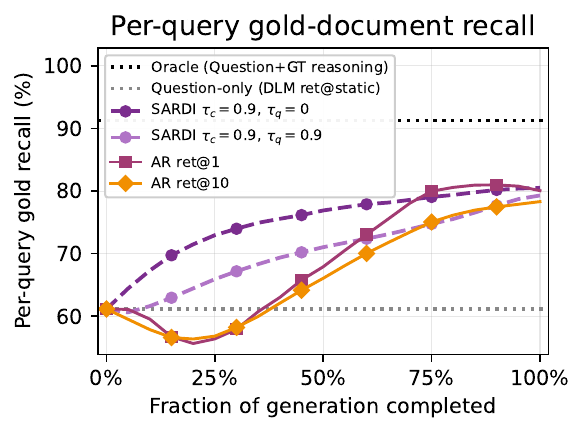}
  \caption{Per-query gold-document recall (fraction of gold passages in the \emph{current} retrieved set) as generation progresses ($K{=}7$, BM25, 2WikiMQA). We restrict to methods that retrieve at every (or every-$N$\textsuperscript{th}) token; agentic methods and FLARE issue specific targeted queries by design, so their per-query recall is low even though cumulative recall is comparable (Appendix~\ref{appendix:cumulative-recall}, \cref{tab:recall-cumulative}).}
  \label{fig:recall-over-time}
\end{figure}

\paragraph{Lookahead surfaces evidence earlier.}
To further analyze the effect of lookahead, \cref{fig:recall-over-time} plots per-query document recall (the fraction of gold passages in the current retrieved set) against generation progress. Two horizontal lines provide baselines: the recall of a question-only query (what static retrieval sees) and the recall of a query built from the ground-truth reasoning trace (roughly the best our framework could reach). As expected, early in generation \ours\ sits well above the AR baselines (a $+19$-point recall gain at $25\%$ of generation) and closes much of the gap toward the oracle line. As expected, the more aggressive the lookahead (lower $\tau_q$), the higher the early recall. By the final step \ours\ converges to AR ret@1, so its main advantage is that strong evidence arrives \emph{early}, letting it make more progress in the opening steps.

\paragraph{Gains concentrate on multi-hop reasoning.}
\Cref{tab:2wiki-qtype-em} breaks 2Wiki down by question type: \ours\ yields over $2.5\times$ EM gains on \emph{inference} and \emph{compositional} questions (which require identifying bridge entities through multi-step reasoning) and leaves single-hop (\emph{comparison}) questions (where static retrieval already suffices) unchanged. 

\begin{table}[h]
  \caption{Question-type breakdown on 2Wiki (EM $\%$). \textsc{SARDI} benefits most from question types requiring multi-step reasoning.}
  \label{tab:2wiki-qtype-em}
  \centering
  \small
  \setlength{\tabcolsep}{4pt}
  \renewcommand{\arraystretch}{1.1}
  \resizebox{\linewidth}{!}{%
    \begin{tabular}{lcccc}
      \toprule
      \textbf{Method} & \multicolumn{1}{c}{\textbf{1-hop}} & \multicolumn{3}{c}{\textbf{Multi-hop}} \\
      \cmidrule(lr){2-2} \cmidrule(lr){3-5}
      & \textbf{comp.} & \textbf{bridge-comp.} & \textbf{compos.} & \textbf{infer.} \\
      \midrule
      \dlms & 85.6 & 65.3 & 16.6 & 14.0 \\
      \dlmours (Ours) $\tau_c{=}0.95$ &
      {84.6\textsubscript{\textcolor{BrickRed}{-1.0}}} &
      {71.7\textsubscript{\textcolor{darkgreen}{+6.4}}} &
      {45.3\textsubscript{\textcolor{darkgreen}{+28.7}}} &
      {37.5\textsubscript{\textcolor{darkgreen}{+23.5}}} \\
      \bottomrule
    \end{tabular}
  }
\end{table}

\subsection{RAG Grounding Promotes Parallel Decoding}
\label{sec:cmi}
\begin{figure}[h]
  \centering
  \includegraphics[width=0.96\columnwidth]{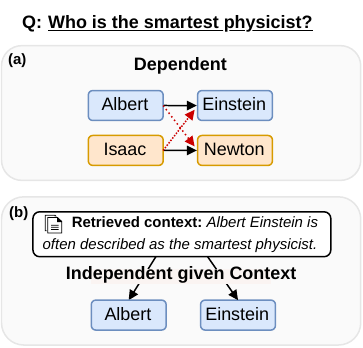}
  \caption{Retrieved evidence decouples adjacent tokens. (a) Without context, the
two tokens are strongly dependent: they must agree to avoid incoherent pairs
such as \emph{Albert Newton}. (b) Given a passage naming the entity, the tokens
are copied from context and become conditionally independent.}
\label{fig:cmi_illustration}
\end{figure}

We argued in \cref{sec:intro} that retrieved evidence reduces inter-token dependence, making RAG well-suited to parallel decoding. We test this by measuring the conditional mutual information (CMI) between adjacent tokens in ground-truth reasoning traces on 2WikiMultiHopQA:
\begin{multline*}
  \mathrm{CMI}(x_i; x_{i+1} \mid D) = \\
  \mathbb{E}_{x_i}\bigl[\mathrm{KL}\bigl(p(x_{i+1} \mid x_i, D)\,\|\,p(x_{i+1} \mid D)\bigr)\bigr].
\end{multline*}
We approximate the expectation over $x_i$ by its top-$7$ values and vary the amount of retrieved evidence in $D$.
\newpage
Intuitively, CMI is high when several completions are plausible and the model has not settled on one. Multi-token names are the clearest case: until the model commits to \emph{which} entity it is naming, the tokens must agree to stay coherent (\emph{Albert Einstein} or \emph{Isaac Newton}, not \emph{Albert Newton}), so fixing one sharply shifts the distribution over the other. Strong evidence removes this coupling: the model copies the entity straight from the retrieved passage instead of coordinating across positions (Figure~\ref{fig:cmi_illustration}).
Formally,
$\mathrm{CMI}(x_i; x_{i+1} \mid D) \leq H(x_{i+1} \mid D)$, so
evidence that sharply reduces the conditional entropy of a position also bounds how much its neighbour can inform it.

\begin{table}[h]
\caption{Conditional mutual information (CMI) between adjacent tokens in \textsc{DREAM-7B} reasoning traces (2WikiMultiHopQA), as gold documents are progressively removed from the context. We separately report entity and non-entity pairs.}
\label{tab:cmi}
  \centering
  \small
  \setlength{\tabcolsep}{6pt}
  \renewcommand{\arraystretch}{1.15}
  \resizebox{0.48\textwidth}{!}{%
    \begin{tabular}{lcccc}
      \toprule
      & \textbf{All gold} & \textbf{Gold$-1$} & \textbf{Gold$-2$} & \textbf{No gold} \\
      \midrule
      CMI (entity)     & 0.060 & 0.219 & 0.396 & {0.588} \\
      CMI (non-entity) & 0.136 & 0.214 & 0.247 & 0.264 \\
      \bottomrule
    \end{tabular}
  }
\end{table}

The results in \Cref{tab:cmi} confirm this. With all gold documents present in the context, entity pairs show very low dependence ($\mathrm{CMI}{=}0.060$), i.e., they can be decoded in parallel. As gold documents are removed, entity-pair CMI rises nearly $10\times$ to $0.588$, while non-entity pairs increase from $0.136$ to $0.264$. Grounding thus removes the inter-token dependence most harmful to parallel decoding: tightly coupled entity spans.
This confirms that the RAG setting is well suited to parallel decoding.

\subsection{Ablations}
\label{sec:ablations}

\paragraph{Retriever choice.}
\ours\ is retriever-agnostic. Replacing BM25 with the E5-base-v2 dense retriever, \ours\ still outperforms the strongest training-free AR baseline (\arone) and stays competitive with the RL-trained Search-R1 (\cref{tab:dense}). Also see more detailed recall results in Appendix~\ref{appendix:cumulative-recall}. Thus, the observed behavior does not depend on the lexical retriever.

\begin{table}[h]
  \caption{\ours\ and strongest baselines with a dense retriever (E5-base-v2). EM ($\times 100$), $K{=}7$.}
  \label{tab:dense}
  \centering
  \small
  \setlength{\tabcolsep}{4pt}
  \renewcommand{\arraystretch}{1.15}
  \resizebox{0.48\textwidth}{!}{%
    \begin{tabular}{lccccc}
      \toprule
      \textbf{Method} & \textbf{2Wiki} & \textbf{Hotpot} & \textbf{CofCA} & \textbf{MuSiQue} & \textbf{SynthW.} \\
      \midrule
      \textcolor{gray}{Search-R1} & \textcolor{gray}{41} & \textcolor{gray}{49} & \textcolor{gray}{44} & \textcolor{gray}{26} & \textcolor{gray}{21} \\
      \arone           & 47 & 47 & 41 & 20 & 22 \\
      \dlmours\ (Ours) & 50 & 52 & 45 & 22 & 23 \\
      \bottomrule
    \end{tabular}
  }
\end{table}

\paragraph{Refresh schedule.}
\ours\ refreshes retrieval at every denoising step, which is wasteful when retrieval is expensive (rerankers, large corpora, tight serving budgets). \Cref{tab:refresh-schedule} varies the refresh frequency: refreshing every $2$ steps costs at most 2 EM. Moreover, we show in Appendix~\ref{appendix:doc-persistence} that the retrieved set is relatively stable: on average $83$--$90\%$ of documents persist between consecutive steps. Both findings suggest that the per-step retrieval cost can be substantially amortized -- a direction we discuss in the Limitations section.
\begin{table}[h]
\caption{\ours\ accuracy under different retrieval refresh frequencies. EM ($\times100$); subscripts show delta to refreshing every step.}  \label{tab:refresh-schedule}
  \centering
  \small
  \setlength{\tabcolsep}{6pt}
  \renewcommand{\arraystretch}{1.15}
  \resizebox{0.48\textwidth}{!}{%
    \begin{tabular}{lcccc}
      \toprule
      \textbf{Refresh} & \textbf{2Wiki} & \textbf{HotpotQA} & \textbf{CofCA} & \textbf{MuSiQue} \\
      \midrule
      every step  & 58 & 48 & 45 & 20 \\
      every 2 & 56\textsubscript{\textcolor{BrickRed}{$-$2}}
              & 47\textsubscript{\textcolor{BrickRed}{$-$1}}
              & 45
              & 20 \\
      every 4 & 52\textsubscript{\textcolor{BrickRed}{$-$6}}
              & 46\textsubscript{\textcolor{BrickRed}{$-$2}}
              & 45
              & 18\textsubscript{\textcolor{BrickRed}{$-$2}} \\
      \bottomrule
    \end{tabular}
  }
\end{table}

\section{Limitations}
\label{sec:limitations}

While \ours\ is training-free in principle, the tested diffusion language models do not yet reliably produce reasoning traces through prompting alone. This is analogous to what \citet{trivedi2023interleaving} observed for early medium-scale autoregressive models, which has since been resolved as AR models matured; we expect the same to happen for DLMs, removing the need for the fine-tuning step.
Additionally, our method refreshes retrieval at every denoising step, which can incur document encoding overhead under naive implementations. Because a large portion of the retrieved documents persists between consecutive steps, integrating Fast-dLLM-style block KV caching~\cite{wu2026fastdllm} is a natural direction for future work. Extending self-augmenting retrieval to latent diffusion language models is another promising direction.

\section{Conclusion}
\label{sec:conclusion}
We introduced \textbf{Self-Augmenting Retrieval for Diffusion Language Models (\ours)}, a training-free, dynamic retrieval framework that exploits two structural opportunities that diffusion language models open up for retrieval-augmented generation. First, the denoising trajectory exposes tentative predictions for the entire response at every step, surfacing salient entities early and turning them into a lookahead signal for retrieval. Second, we have shown that grounding generation in retrieved evidence sharply reduces inter-token dependence, making RAG a regime especially well suited to parallel decoding. \ours\ leverages both: it interleaves retrieval with denoising, refining the query as the response takes shape. On five multi-hop QA benchmarks, \ours\ substantially improves over static retrieval, matches or beats every training-free AR baseline, and runs up to $4\times$ faster. More broadly, denoising trajectories offer a natural signal for dynamic retrieval, and we expect \ours\ to strengthen as diffusion language models mature.

\section*{Acknowledgements}
We thank Alexander Kreibich and Lucas He for productive discussions that helped motivate this work. JL is supported by a Google PhD Fellowship, LZ by a LinkedIn PhD Fellowship, and DG by an Empire AI Postdoctoral Fellowship. We are grateful to NVIDIA for access to the DGX Station compute platform through their Early Access Program. This work was further supported by the National Science Foundation (NSF) under grants OAC-2118310 and IIS-2530143, through the AI Research Institutes program (Award No.\ DMR-2433348), by funding from NewYork-Presbyterian for the NYP-Cornell Cardiovascular AI Collaboration, and by arXiv. Finally, we thank the anonymous reviewers, whose feedback strengthened the paper by motivating the analyses of early-generation recall and mutual information.

\section*{Impact Statement}

This work studies retrieval-augmented generation for diffusion language models.
Improving evidence grounding can reduce hallucinations in knowledge-intensive settings, but the same retrieval mechanisms could produce more convincing misinformation if paired with untrusted corpora.
We recommend deploying \ours\ with curated sources and provenance tracking, such as citing retrieved passages alongside generated answers.
All experiments use publicly available QA benchmarks and open retrieval corpora.
\newpage

\bibliography{main}
\bibliographystyle{icml2026}

\newpage
\appendix
\onecolumn

\section{Fine-Tuning Analysis}
\label{appendix:sft-analysis}

\begin{table}[h]
  \caption{Output-mode breakdown for base and fine-tuned \textsc{DREAM-7B} and Qwen2.5-7B on 2WikiMultiHopQA (static $K{=}7$ RAG, $n{=}500$).}
  \label{tab:sft-failure}
  \centering
  \small
  \setlength{\tabcolsep}{4pt}
  \renewcommand{\arraystretch}{1.15}
  \resizebox{0.85\textwidth}{!}{%
    \begin{tabular}{lcccc}
      \toprule
       & \textbf{DREAM-7B} & \textbf{Qwen2.5-7B} & \textbf{DREAM-7B (SFT)} & \textbf{Qwen2.5-7B (SFT)} \\
      \midrule
      Correct (with reasoning trace) & $1\%$  & $32\%$ & $\mathbf{44\%}$ & $\mathbf{44\%}$ \\
      Correct (no reasoning trace)   & $27\%$ & $0\%$  & $0\%$  & $0\%$ \\
      Wrong                          & $72\%$ & $68\%$ & $56\%$ & $56\%$ \\
      \bottomrule
    \end{tabular}
  }
\end{table}

To motivate the fine-tuning stage and isolate its effect from the retrieval mechanism, we present an output analysis of base and fine-tuned models on $500$ examples from 2WikiMultiHopQA under static $K{=}7$ retrieval.
\Cref{tab:sft-failure} reports the proportion of outputs in three categories: correct with a structured reasoning trace, correct without a reasoning trace, and wrong (which folds in format errors such as missing the \texttt{\#\#\#} answer marker).

Two observations are worth highlighting:
\begin{enumerate}
    \item \textbf{Base \textsc{DREAM-7B} cannot produce structured reasoning from prompting alone.} Only $1\%$ of base \textsc{DREAM-7B} outputs include a correct reasoning trace. The same-sized AR Qwen2.5-7B reasons correctly in $32\%$ of cases. \ours\ relies on intermediate states surfacing salient entities in the \emph{reasoning trace}, so this failure mode makes \ours\ inapplicable without fine-tuning.
    \item \textbf{Fine-tuning equalizes capability.} On this $n{=}500$ subset both models land at $44\%$ correct-with-reasoning; on the full test set they sit within 1~EM of each other ($\dlms{=}43.7\%$, $\ars{=}44.5\%$; \cref{tab:main-results}). Any subsequent gap between \ours\ and matched-fine-tuning AR baselines is therefore attributable to the retrieval mechanism rather than the underlying model.
\end{enumerate}

This is analogous to what~\citet{trivedi2023interleaving} observed for early small AR models: ``IRCoT relies on the base LM to have a zero or few-shot CoT-generation ability. While this is commonly available in large LMs (over 100B), it's not as common for small LMs (under 20B)\dots smaller LMs will likely increasingly acquire such ability.''
That prediction proved correct for AR models; we expect the same for DLMs.

\subsection{Fine-Tuning Data}
\label{appendix:data_construction}

We construct supervised fine-tuning data from the training splits of 2WikiMultiHopQA~\cite{ho-etal-2020-2wiki} and HotpotQA~\cite{yang2018hotpotqa}, augmented with synthetically generated chain-of-thought reasoning traces.

\paragraph{Synthetic reasoning traces.}
For each training example, we prompt \emph{gpt-4o-mini} with the question $q$, gold answer $a$, and supporting documents to generate a step-by-step reasoning trace that concludes with the final answer in the format \texttt{\#\#\# [answer]}.

\paragraph{Example.}
The following illustrates a training example from 2WikiMultiHopQA:

\begin{verbatim}
Question: Where was the director of film Ronnie Rocket born?

Reasoning (synthetically generated):
Step 1: Ronnie Rocket is directed by David Lynch.
Step 2: David Lynch was born in Missoula, Montana.

### Montana
\end{verbatim}

\paragraph{Training context.}
During training, we present each training example together with its gold documents, reinforcing Dream-7B to generate reasoning traces when solving RAG problems.

\subsection{Fine-Tuning Configuration}
\label{appendix:finetuning_config}

We fine-tune \textsc{DREAM-7B} using fully sharded data parallel (FSDP) on 2$\times$ NVIDIA B200 GPUs.
\Cref{tab:hyperparameters} summarizes the hyperparameters.
To ensure a fair comparison, we apply the same configuration and training data to all autoregressive baselines (Qwen2.5-7B), isolating differences attributable to the generation paradigm.

\begin{table}[h]
\centering
\caption{Fine-tuning hyperparameters.}
\label{tab:hyperparameters}
\begin{tabular}{ll}
\toprule
\textbf{Hyperparameter} & \textbf{Value} \\
\midrule
Base model & DREAM-7B \\
Training epochs & 3 \\
Learning rate & $2 \times 10^{-6}$ \\
Global batch size & 256 \\
Micro batch size per GPU & 16 \\
Maximum sequence length & 2048 tokens \\
Optimizer & AdamW \\
Hardware & 2$\times$ NVIDIA B200 GPUs \\
\bottomrule
\end{tabular}
\end{table}

\section{Prompt Templates}
\label{appendix:prompts}

\subsection{RAG Input Format}
\label{appendix:rag_prompt}

All models receive input in the following format:

\begin{verbatim}
Use ONLY the provided facts to answer the question.
Think step-by-step, then provide the final answer after the "###" marker.

Question:
{question}

Facts:
{facts}
\end{verbatim}

\noindent where \texttt{\{question\}} is the input question and \texttt{\{facts\}} contains the concatenated retrieved passages.

\subsection{Expected Output Format}
\label{appendix:output_format}

Models are trained to produce outputs in the following format:

\begin{verbatim}
Step 1: [First reasoning step grounded in documents]
Step 2: [Second reasoning step]
...
### [final answer]
\end{verbatim}

\noindent This format serves two purposes: it enables extraction of intermediate entities from partial reasoning traces for self-augmenting retrieval, and it provides a consistent extraction point (\texttt{\#\#\#}) for evaluation.

\section{Additional Experimental Details}
\label{appendix:experimental_details}

\paragraph{Evaluation protocol.}
We extract the final answer by parsing text following the \texttt{\#\#\#} marker.
Exact Match (EM) is computed after normalizing both predicted and gold answers and checking for string equality.

\paragraph{Retrieval configuration.}
We use BM25~\cite{bm25} for all experiments unless otherwise specified.
For the dense-retriever results in~\cref{sec:ablations} we use E5-base-v2~\citep{wang2022_e5}.
For \ours, retrieval is performed at every denoising iteration using the concatenation of the original question and the current intermediate response as the query, with $K{=}7$ passages per iteration unless otherwise specified.

\paragraph{Search-R1 setup.}
\label{appendix:searchr1-setup}
The four AR baselines (\ars, \arten, \arone, \ara) and the training-free agentic baselines (AdaptiveRAG, ReAct) all share the same Qwen2.5-7B backbone, supervised-fine-tuned on the combined training sets of 2WikiMultiHopQA and HotpotQA (Appendix~\ref{appendix:sft-analysis}); any accuracy gap with \ours\ therefore isolates the retrieval mechanism.
Search-R1 is the only baseline that does \emph{not} share this backbone: we evaluate the authors' publicly released checkpoint, which they trained via PPO on the merged HotpotQA and NQ training sets using a multi-stage reward~\citep{jin2025searchr1}.
Reproducing this RL training pipeline was not feasible within our compute budget, so we accept this additional confound and treat Search-R1 as an RL-trained reference point rather than a controlled comparison (hence the gray styling in~\cref{tab:main-results}).

\section{Per-Method Recall: Full Tables}
\label{appendix:cumulative-recall}

\Cref{fig:recall-over-time} in the main text reports gold-document recall over generation progress for SARDI and the AR retrieval baselines that fire at every (or every-$N$\textsuperscript{th}) token.
For completeness, this section reports the full per-method numbers, including the agentic methods (AdaptiveRAG, ReAct, Search-R1) and AR speculative-lookahead (FLARE).

We report two complementary recall metrics:
\begin{itemize}[leftmargin=*, itemsep=2pt, topsep=0pt]
    \item \textbf{Per-query final recall} (\cref{tab:recall-full}): fraction of gold documents present in the retrieved set $D$ at the \emph{final} retrieval step, averaged over questions. This is the metric most directly tied to answer accuracy on a single forward pass.
    \item \textbf{Cumulative recall} (\cref{tab:recall-cumulative}): fraction of gold documents that have appeared in $D$ at \emph{any} point during generation, i.e., the union over all retrieval steps. This view is fairer to methods that issue few but targeted queries.
\end{itemize}

\textbf{Caveat: cumulative recall depends on retrieval volume.}
A method that issues many queries (e.g., \ours, AR ret@1) trivially accumulates more unique documents than one that issues few targeted queries (e.g., Search-R1), and cumulative recall therefore is \emph{not} directly comparable across methods without controlling for total documents touched.
\Cref{tab:recall-cumulative} accordingly reports the average number of unique documents retrieved per question as a subscript next to each cumulative-recall value, so the reader can normalize.

\begin{table*}[h]
  \caption{Per-query gold-document recall at the final retrieval step ($\times 100$). Methods that issue specific targeted queries (AdaptiveRAG, ReAct, AR ret@adaptive/FLARE, Search-R1) retrieve few documents per query, so per-query recall is low by design even when their cumulative coverage (\cref{tab:recall-cumulative}) is comparable. \ours\ matches AR ret@1 across $K$ and retrievers.}
  \label{tab:recall-full}
  \centering
  \small
  \setlength{\tabcolsep}{4pt}
  \renewcommand{\arraystretch}{1.05}
  \resizebox{\textwidth}{!}{%
    \begin{tabular}{lcccccccccccccccc}
      \toprule
      & \multicolumn{8}{c}{\textbf{2Wiki}} & \multicolumn{8}{c}{\textbf{MuSiQue}} \\
      \cmidrule(lr){2-9} \cmidrule(lr){10-17}
      & \multicolumn{4}{c}{\textit{BM25}} & \multicolumn{4}{c}{\textit{Dense}} & \multicolumn{4}{c}{\textit{BM25}} & \multicolumn{4}{c}{\textit{Dense}} \\
      \cmidrule(lr){2-5} \cmidrule(lr){6-9} \cmidrule(lr){10-13} \cmidrule(lr){14-17}
      \textbf{Method} & $K{=}3$ & $K{=}7$ & $K{=}10$ & $K{=}15$ & $K{=}3$ & $K{=}7$ & $K{=}10$ & $K{=}15$ & $K{=}3$ & $K{=}7$ & $K{=}10$ & $K{=}15$ & $K{=}3$ & $K{=}7$ & $K{=}10$ & $K{=}15$ \\
      \midrule
      AdaptiveRAG & 5 & 5 & 4 & 3 & -- & 5 & 4 & 3 & 9 & 10 & 9 & 7 & -- & 9 & -- & -- \\
      ReAct & 38 & 48 & 51 & 54 & 27 & 31 & 31 & 33 & 26 & 34 & 37 & 40 & 27 & 31 & 32 & 35 \\
      AR ret@adaptive (FLARE) & 22 & 27 & 29 & 32 & 14 & 17 & 18 & 19 & 14 & 20 & 22 & 25 & 14 & 20 & 21 & 22 \\
      AR ret@10 & 65 & 78 & 82 & 86 & 54 & 61 & 62 & 63 & 42 & 53 & 56 & 61 & 38 & 47 & 51 & 54 \\
      AR ret@1 & 66 & 80 & 85 & 88 & 54 & 61 & 63 & 64 & 43 & 54 & 58 & 62 & 38 & 48 & 51 & 54 \\
      Search-R1 & 42 & 53 & 57 & 62 & 28 & 31 & 33 & 35 & 29 & 38 & 42 & 45 & 28 & 35 & 37 & 41 \\
      \textbf{\ours\ $\tau_c{=}0.9$} & 66 & 81 & 83 & 86 & 55 & 63 & 62 & 63 & 43 & 55 & 56 & 60 & 39 & 50 & 53 & 55 \\
      \textbf{\ours\ $\tau_c{=}0.95$} & 67 & 81 & 84 & 88 & 56 & 63 & 64 & 65 & 44 & 56 & 59 & 64 & 40 & 50 & 53 & 56 \\
      \bottomrule
    \end{tabular}%
  }
\end{table*}

\begin{table*}[t]
  \caption{Cumulative gold-document recall ($\times 100$, main number) and average number of unique documents retrieved per question (\textsubscript{subscript}). Cumulative recall measures the union of all gold documents surfaced at any point during generation, but is \emph{not directly comparable across methods that retrieve different total document counts}: a method that issues many queries (e.g., \ours, AR ret@1) trivially accumulates more docs than one that issues few (e.g., Search-R1). The subscripts let the reader normalize. Even controlling for total docs, \ours\ remains competitive with the strongest baselines.}
  \label{tab:recall-cumulative}
  \centering
  \small
  \setlength{\tabcolsep}{4pt}
  \renewcommand{\arraystretch}{1.05}
  \resizebox{\textwidth}{!}{%
    \begin{tabular}{lcccccccccccccccc}
      \toprule
      & \multicolumn{8}{c}{\textbf{2Wiki}} & \multicolumn{8}{c}{\textbf{MuSiQue}} \\
      \cmidrule(lr){2-9} \cmidrule(lr){10-17}
      & \multicolumn{4}{c}{\textit{BM25}} & \multicolumn{4}{c}{\textit{Dense}} & \multicolumn{4}{c}{\textit{BM25}} & \multicolumn{4}{c}{\textit{Dense}} \\
      \cmidrule(lr){2-5} \cmidrule(lr){6-9} \cmidrule(lr){10-13} \cmidrule(lr){14-17}
      \textbf{Method} & $K{=}3$ & $K{=}7$ & $K{=}10$ & $K{=}15$ & $K{=}3$ & $K{=}7$ & $K{=}10$ & $K{=}15$ & $K{=}3$ & $K{=}7$ & $K{=}10$ & $K{=}15$ & $K{=}3$ & $K{=}7$ & $K{=}10$ & $K{=}15$ \\
      \midrule
      AdaptiveRAG & 71\textsubscript{6} & 79\textsubscript{13} & 77\textsubscript{14} & 62\textsubscript{15} & -- & -- & -- & -- & 48\textsubscript{7} & 56\textsubscript{13} & 55\textsubscript{14} & 48\textsubscript{15} & -- & -- & -- & -- \\
      ReAct & 73\textsubscript{7} & 81\textsubscript{15} & 82\textsubscript{21} & 83\textsubscript{31} & 57\textsubscript{7} & 57\textsubscript{15} & 57\textsubscript{21} & 58\textsubscript{31} & 50\textsubscript{6} & 60\textsubscript{15} & 63\textsubscript{22} & 65\textsubscript{33} & 54\textsubscript{7} & 60\textsubscript{16} & 62\textsubscript{22} & 64\textsubscript{33} \\
      AR ret@adaptive (FLARE) & 73\textsubscript{8} & 82\textsubscript{20} & 84\textsubscript{29} & 87\textsubscript{43} & 63\textsubscript{9} & 66\textsubscript{22} & 67\textsubscript{32} & 68\textsubscript{49} & 51\textsubscript{9} & 62\textsubscript{22} & 65\textsubscript{31} & 69\textsubscript{48} & 52\textsubscript{9} & 60\textsubscript{22} & 64\textsubscript{32} & 65\textsubscript{48} \\
      AR ret@10 & 71\textsubscript{6} & 83\textsubscript{15} & 86\textsubscript{22} & 89\textsubscript{34} & 61\textsubscript{6} & 66\textsubscript{15} & 67\textsubscript{22} & 68\textsubscript{33} & 49\textsubscript{6} & 60\textsubscript{16} & 64\textsubscript{23} & 68\textsubscript{34} & 46\textsubscript{6} & 55\textsubscript{15} & 58\textsubscript{22} & 62\textsubscript{32} \\
      AR ret@1 & 73\textsubscript{7} & 85\textsubscript{20} & 89\textsubscript{29} & 91\textsubscript{46} & 64\textsubscript{9} & 68\textsubscript{23} & 69\textsubscript{33} & 70\textsubscript{50} & 52\textsubscript{8} & 63\textsubscript{21} & 67\textsubscript{30} & 71\textsubscript{45} & 50\textsubscript{9} & 59\textsubscript{22} & 62\textsubscript{31} & 65\textsubscript{46} \\
      Search-R1 & 74\textsubscript{6} & 85\textsubscript{16} & 88\textsubscript{23} & 90\textsubscript{34} & 60\textsubscript{7} & 63\textsubscript{17} & 63\textsubscript{25} & 63\textsubscript{37} & 57\textsubscript{7} & 65\textsubscript{16} & 68\textsubscript{23} & -- & 59\textsubscript{7} & 65\textsubscript{17} & 67\textsubscript{23} & 70\textsubscript{34} \\
      \textbf{\ours\ $\tau_c{=}0.9$} & 70\textsubscript{5} & 83\textsubscript{14} & 85\textsubscript{20} & 88\textsubscript{31} & 61\textsubscript{6} & 68\textsubscript{17} & 66\textsubscript{24} & 67\textsubscript{37} & 47\textsubscript{7} & 63\textsubscript{19} & 61\textsubscript{28} & 65\textsubscript{44} & 47\textsubscript{7} & 58\textsubscript{19} & 59\textsubscript{28} & 61\textsubscript{42} \\
      \textbf{\ours\ $\tau_c{=}0.95$} & 71\textsubscript{5} & 83\textsubscript{14} & -- & 90\textsubscript{33} & 63\textsubscript{7} & 68\textsubscript{18} & 70\textsubscript{26} & 71\textsubscript{40} & 52\textsubscript{7} & 64\textsubscript{20} & 68\textsubscript{30} & 73\textsubscript{46} & 49\textsubscript{8} & 59\textsubscript{20} & 62\textsubscript{29} & 66\textsubscript{44} \\
      \bottomrule
    \end{tabular}%
  }
\end{table*}

\section{Document Persistence Across Refresh Steps}
\label{appendix:doc-persistence}

\Cref{tab:refresh-schedule} relies on the observation that retrieved documents change only gradually between consecutive denoising steps, so document-level KV-cache reuse is feasible.
We measure the average overlap between consecutive retrieved sets as $|D^t \cap D^{t-1}| / |D^{t-1}|$:

\begin{center}
\begin{tabular}{lc}
\toprule
\textbf{Dataset} & \textbf{\% docs retained per step} \\
\midrule
2WikiMultiHopQA & $88\%$ \\
HotpotQA & $83\%$ \\
CofCA & $89\%$ \\
MuSiQue & $84\%$ \\
SynthWorlds-SM & $90\%$ \\
\bottomrule
\end{tabular}
\end{center}

Since $83$--$90\%$ of documents persist between consecutive steps, only a small fraction needs recomputation, making document-level KV caching similar to Fast-dLLM~\citep{wu2026fastdllm} an attractive direction for combining \ours\ with diffusion-decoding-acceleration techniques.

\section{Accuracy vs.\ Throughput: Full Datasets}
\label{appendix:accuracy_vs_qps}

\begin{figure*}[h]
  \centering
  \includegraphics[width=0.8\textwidth]{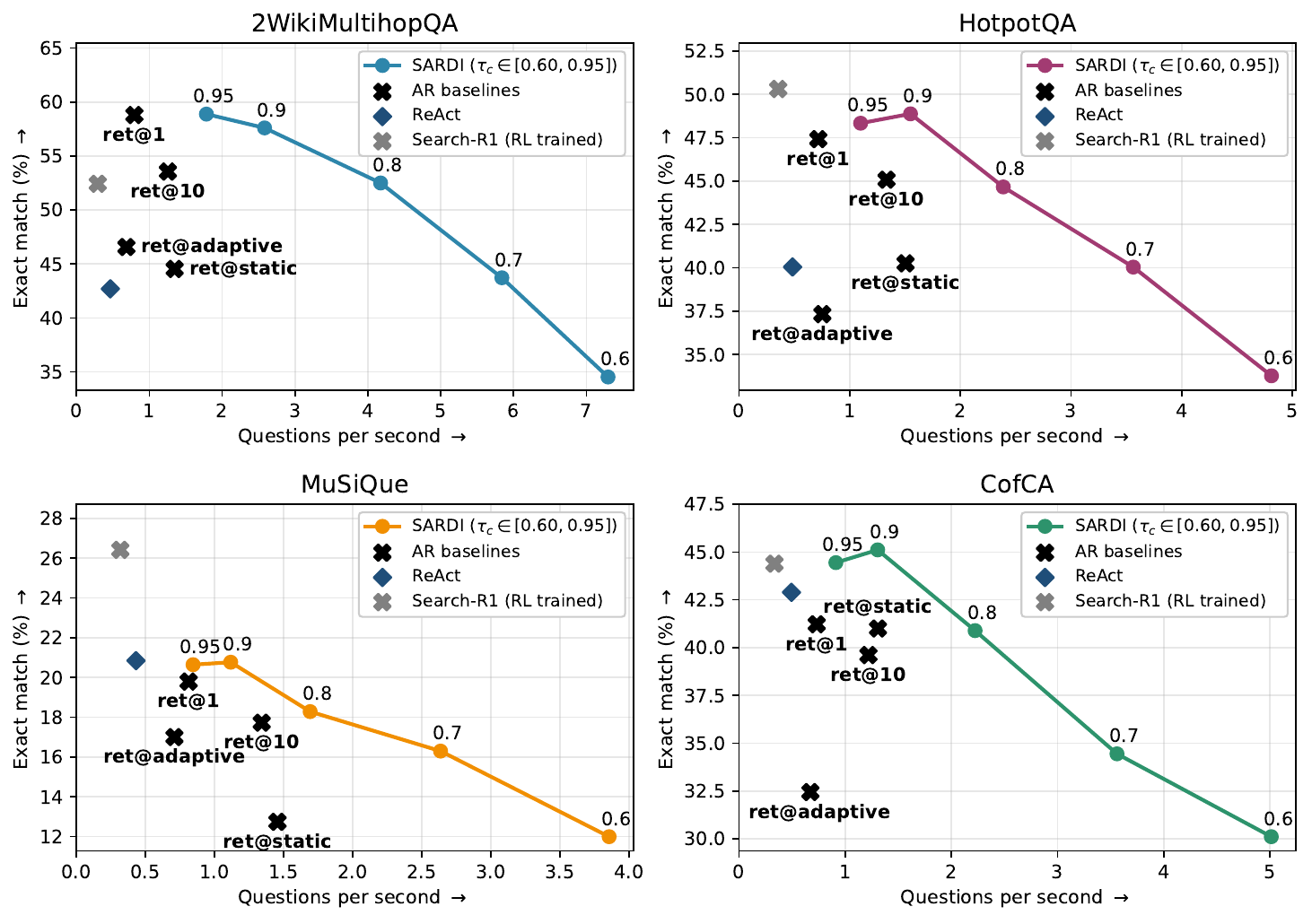}
  \caption{Accuracy vs.\ throughput trade-off across four benchmarks, companion to \cref{fig:accuracy_vs_qps}.}
  \label{fig:accuracy_vs_qps-appendix}
\end{figure*}

\Cref{fig:accuracy_vs_qps} in the main text shows the Pareto frontier on 2WikiMultiHopQA and CofCA. \Cref{fig:accuracy_vs_qps-appendix} reports the same plot across all five benchmarks. \ours\ dominates the autoregressive iterative-retrieval baselines on every dataset and matches Search-R1 at $4$--$9\times$ lower latency.

\section{Threshold-Based vs.\ Fixed-Step Decoding}
\label{appendix:threshold-decoding}

\begin{table}[h]
  \caption{Threshold-based unmasking matches fixed-step accuracy at up to $3\times$ the speed. EM ($\times 100$) and time per example (s).}
  \label{tab:threshold-ablation}
  \centering
  \small
  \setlength{\tabcolsep}{4pt}
  \renewcommand{\arraystretch}{1.15}
  \resizebox{0.48\textwidth}{!}{%
    \begin{tabular}{lcccc}
      \toprule
      & \multicolumn{1}{c}{\textbf{2Wiki}}
      & \multicolumn{1}{c}{\textbf{HotpotQA}}
      & \multicolumn{1}{c}{\textbf{CofCA}}
      & \multicolumn{1}{c}{\textbf{MuSiQue}} \\
      & EM / T & EM / T & EM / T & EM / T \\
      \midrule
      50 fixed steps & 59 / 1.33 & 47 / 1.46 & 43 / 1.48 & 20 / 1.42 \\
      $\tau_c{=}0.9$
        & 58\textsubscript{\textcolor{BrickRed}{$-$1}} / 0.39\textsubscript{\textcolor{darkgreen}{$-$0.94}}
        & 48\textsubscript{\textcolor{darkgreen}{$+$1}} / 0.64\textsubscript{\textcolor{darkgreen}{$-$0.82}}
        & 45\textsubscript{\textcolor{darkgreen}{$+$2}} / 0.75\textsubscript{\textcolor{darkgreen}{$-$0.73}}
        & 20 / 0.88\textsubscript{\textcolor{darkgreen}{$-$0.54}} \\
      $\tau_c{=}0.95$
        & 59 / 0.56\textsubscript{\textcolor{darkgreen}{$-$0.77}}
        & 49\textsubscript{\textcolor{darkgreen}{$+$2}} / 0.90\textsubscript{\textcolor{darkgreen}{$-$0.56}}
        & 45\textsubscript{\textcolor{darkgreen}{$+$2}} / 1.09\textsubscript{\textcolor{darkgreen}{$-$0.39}}
        & 21\textsubscript{\textcolor{darkgreen}{$+$1}} / 1.19\textsubscript{\textcolor{darkgreen}{$-$0.23}} \\
      \bottomrule
    \end{tabular}
  }
\end{table}

\Cref{tab:threshold-ablation} compares fixed-step decoding ($T{=}50$) against threshold-based unmasking at the two commit thresholds used in the main paper, $\tau_c{=}0.9$ and $\tau_c{=}0.95$.
Threshold decoding matches fixed-step accuracy at up to $3\times$ the speed.

\section{Query-Threshold Sweep: Full Numbers}
\label{appendix:qct-sweep}

\Cref{fig:qct-sweep} in the main text plots the $\tau_q$ sweep on 2WikiMultiHopQA. \Cref{tab:qct-sweep} reports full EM numbers across all four benchmarks for which $\tau_q$ sweeps were run, at both commit thresholds $\tau_c \in \{0.9, 0.95\}$ and ten coarse $\tau_q$ values spanning the spectrum from aggressive lookahead ($\tau_q{=}0$, the default \ours\ setting) to the conservative endpoint ($\tau_q{=}0.9$, querying only on near-committed tokens). The pattern observed on 2WikiMultiHopQA reproduces on HotpotQA, MuSiQue, and SynthWorlds-SM: peak accuracy lands at $\tau_q \in [0, 0.1]$ on every (dataset, $\tau_c$) cell, and EM generally declines as the query threshold tightens. The conservative endpoint loses $4$--$6$ EM relative to the default, consistent with our hypothesis that retrieval benefits from speculative future tokens that are not yet stable enough to commit.

\begin{table*}[t]
  \caption{Query-threshold sweep, full numbers. EM (\%) on the four datasets with $\tau_q$ sweeps, at commit thresholds $\tau_c \in \{0.9, 0.95\}$.}
  \label{tab:qct-sweep}
  \centering
  \small
  \setlength{\tabcolsep}{4pt}
  \renewcommand{\arraystretch}{1.1}
  \resizebox{\textwidth}{!}{%
    \begin{tabular}{llcccccccccc}
      \toprule
      \textbf{Dataset} & $\tau_c$ & $\tau_q{=}0$ & $\tau_q{=}0.1$ & $\tau_q{=}0.2$ & $\tau_q{=}0.3$ & $\tau_q{=}0.4$ & $\tau_q{=}0.5$ & $\tau_q{=}0.6$ & $\tau_q{=}0.7$ & $\tau_q{=}0.8$ & $\tau_q{=}0.9$ \\
      \midrule
      \multirow{2}{*}{2WikiMultihopQA} & 0.9 & 57.8 & 58.1 & 57.4 & 57.3 & 56.6 & 55.6 & 54.7 & 54.5 & 54.0 & 53.2 \\
       & 0.95 & 59.1 & 59.4 & 58.7 & 58.4 & 57.6 & 57.1 & 56.2 & 55.6 & 54.8 & 54.0 \\
      \midrule
      \multirow{2}{*}{HotpotQA} & 0.9 & 48.5 & 48.1 & 47.6 & 46.4 & 45.9 & 46.0 & 44.8 & 44.7 & 44.4 & 44.1 \\
       & 0.95 & 48.7 & 48.4 & 47.8 & 47.0 & 46.3 & 45.6 & 45.3 & 44.9 & 44.4 & 44.1 \\
      \midrule
      \multirow{2}{*}{MuSiQue} & 0.9 & 20.5 & 20.6 & 19.7 & 19.2 & 18.2 & 16.7 & 16.3 & 15.7 & 15.7 & 15.2 \\
       & 0.95 & 20.6 & 20.6 & 19.5 & 18.9 & 17.9 & 17.1 & 16.3 & 16.3 & 15.6 & 14.7 \\
      \midrule
      \multirow{2}{*}{SynthWorlds} & 0.9 & 21.1 & 21.4 & 21.1 & 19.9 & 17.9 & 17.5 & 17.9 & 17.0 & 16.8 & 17.2 \\
       & 0.95 & 21.7 & 21.5 & 21.3 & 19.8 & 19.5 & 18.2 & 17.5 & 17.5 & 17.0 & 16.9 \\
      \bottomrule
    \end{tabular}%
  }
\end{table*}

\section{Qualitative Example: Threshold-Based Unmasking}
\label{appendix:qualitative}

A central hypothesis of this work is that RAG exhibits a structure uniquely amenable to parallel decoding: when retrieved evidence is sufficiently informative, many output tokens become conditionally independent given the evidence and can be committed simultaneously.
Threshold-based unmasking (\cref{eq:unmask-set}) exploits this by committing all high-confidence tokens at once rather than artificially limiting parallelism.
The following example illustrates this behavior for $\tau_c{=}0.8$, requiring only 2 steps:

{\small
\emph{Where was the place of death of the director of Fight Of The Tertia?}

$x_2=$ ``\texttt{[M][M][M][M][M][M][M][M][M][M][M][M][M]}''

$D_2=$ ``Fight of the Tertia is a 1952 West German family film directed by Erik Ode [\dots]''

$x_1=$ ``\texttt{(1)\ Fight of the Tertia is directed by Erik Ode.\ (2)\ Erik Ode died in [M][M][M]}''

$D_1=$ ``Erik Ode (born Fritz Erik Signy Odemar, 6 November 1910 in Berlin, died 19 July 1983 in Kreuth-Wei\ss ach) was a [\dots]''

$x_0=$ ``\texttt{(1)\ Fight of the Tertia is directed by Erik Ode.\ (2)\ Erik Ode died in Kreuth. \#\#\# Kreuth}''
}



\end{document}